\documentclass[10pt,twocolumn,letterpaper]{article}

\usepackage{iccv}
\usepackage{times}
\usepackage{epsfig}
\usepackage{graphicx}
\usepackage{amsmath}
\usepackage{amssymb}
\usepackage[square,numbers]{natbib}
\usepackage{booktabs}
\usepackage{subcaption}
\usepackage{pifont}
\usepackage{siunitx}

\usepackage[pagebackref=true,breaklinks=true,letterpaper=true,colorlinks,bookmarks=false]{hyperref}

\iccvfinalcopy 

\def\iccvPaperID{9367} 

\ificcvfinal\fi

\begin{document}
\captionsetup[table]{position=below}

\title{The Untapped Potential of Off-the-Shelf Convolutional Neural Networks}

\author{Matthew Inkawhich\textsuperscript{1}\quad Nathan Inkawhich\textsuperscript{1}\quad Eric Davis\textsuperscript{2}\quad Hai Li\textsuperscript{1}\quad Yiran Chen\textsuperscript{1}\\
\textsuperscript{1}Duke University\quad \textsuperscript{2}SRC Inc.\\
{\tt\small matthew.inkawhich@duke.edu}
}

\maketitle
\ificcvfinal\thispagestyle{empty}\fi

\begin{abstract}
Over recent years, a myriad of novel convolutional network architectures have been developed to advance state-of-the-art performance on challenging recognition tasks. As computational resources improve, a great deal of effort has been placed in efficiently scaling up existing designs and generating new architectures with Neural Architecture Search (NAS) algorithms. While network topology has proven to be a critical factor for model performance, we show that significant gains are being left on the table by keeping topology static at inference-time. Due to challenges such as scale variation, we should not expect static models configured to perform well across a training dataset to be optimally configured to handle all test data. In this work, we seek to expose the exciting potential of inference-time-dynamic models. By allowing just four layers to dynamically change configuration at inference-time, we show that existing off-the-shelf models like ResNet-50 are capable of over 95\% accuracy on ImageNet. This level of performance currently exceeds that of models with over 20x more parameters and significantly more complex training procedures.
\end{abstract}

\vspace{-3mm}
\section{Introduction}

\begin{figure}[t]
    \centering
    \includegraphics[width=.85\columnwidth]{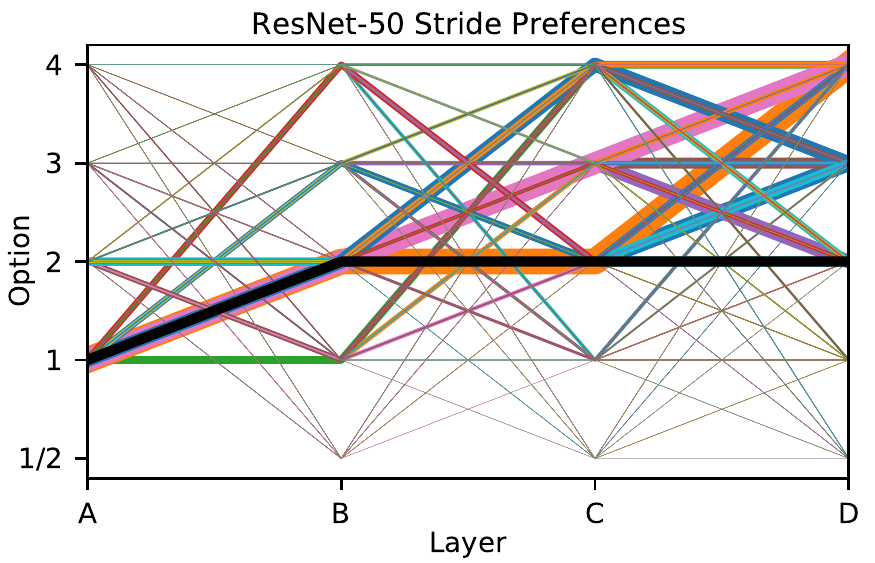}
    \vspace{-3mm}
    \caption{\textbf{Preferred strides of pre-trained ResNet-50 model.} This visualization shows stride preferences of an off-the-shelf ResNet-50 model. Each colored line stretches from layer A to D and represents a different possible stride permutation over the depth of the model. Line thickness encodes the proportion of validation samples that most prefer that permutation of strides. The default stride permutation for layers A, B, C, and D is $S$=(1, 2, 2, 2), which is represented by the black line.}
    \label{fig:best_paths}
    \vspace{-5mm}
\end{figure}

The evolution of Convolutional Neural Network (CNN) design has yielded tremendous advances in performance for many vision tasks. Canonical designs such as LeNet \cite{lenet} and AlexNet \cite{alexnet} established a basic set of design patterns and that have inspired countless new architectures. Subsequent designs such as VGGNet \cite{vgg}, ResNet \cite{resnet}, and GoogLeNet \cite{inception_v1} have introduced new heuristics to make models deeper and more capable. As compute resources advance, there has been a shift in effort away from designing new hand-crafted architectures, and into scaling up existing designs \cite{efficientnet} and leveraging Neural Architecture Search (NAS) \cite{NAS_zoph, NAS_baker} to learn optimal topologies.

In this work, we argue that regardless of how innovative the architectural design of a CNN is, the performance may be inherently limited if it is kept static at inference time. We define a static network as one whose topology is the same for all test samples, and a dynamic network as one whose topology is free to change from sample to sample. Our results show that if off-the-shelf benchmark models such as ResNet-50 \cite{resnet} are made dynamic at inference-time, they are capable of accuracy levels comparable to models with over 20x more parameters \textit{and} efficiency levels comparable to models with 7x fewer parameters. To make a network dynamic we allow attributes of the convolutional layers that are often considered ``set-and-forget" to change from sample to sample. These attributes include kernel stride, dilation, and size. Note that we implement these dynamic attributes such that the parameter tensors themselves, which are pre-trained and downloaded from the PyTorch Model Zoo \cite{pytorch}, are not altered. Rather, the application of the kernel on the input features during convolution is changed.

One core challenge that static models are ill-poised for but dynamic models excel at handling is feature scale variation. In this context, feature scale variation refers to the disparity in the size and shape of image details in pixel-space that are processed into feature space. In natural imagery such as the data in ImageNet \citep{imagenet}, scale variation is a prominent challenge. Not only do the physical manifestations of objects of different classes vary (\eg, grasshopper vs. African elephant), but the object depth and orientation also vary dramatically from image to image. Despite the fact that this scale variation is present in the training set and accounted for by most static CNNs via data augmentation, we opine that the static convolutional feature extraction pipeline is NOT optimal for all samples during inference and thus critically limits the upper bound of expected generalization. Our hypothesis is that off-the-shelf models have the ability to extract meaningful features from more samples than they are given credit for. The fundamental feature patterns that the learned kernels activate on do not change when we alter parameters such as stride, dilation, and kernel size, but the scale of the features that activate them do. For some samples, using larger kernel strides than the default may be what is required for the model to properly extract pertinent features. Figure \ref{fig:best_paths} is a visualization of stride preferences for a pre-trained ResNet-50 model. Each different colored line represents a unique permutation of strides over four dynamic layers, and its width encodes the proportion of validation samples that are most correctly predicted with that permutation. Interestingly, the preferences are quite uniformly distributed and the default permutation (the black line) is not even the most commonly preferred!

Note that the goal of this work is not to propose an implementation for a truly dynamic model. Rather, our goal is to perform an in-depth analysis of the performance that inference-time-dynamic (ITD) models can achieve if implemented well. In other words, we are interested in the performance of dynamic model oracles that always choose the best (or worst) configuration for each sample. We do this by using a comprehensive evaluation process and a series of novel analytical techniques. We also leverage this evaluation process to help interpret model behaviors, and find interesting inconsistencies with several familiar design heuristics. Overall, we hope that this work will inspire important future research that further investigates and exploits this significant untapped potential of CNNs.
\section{Related work}

The scale variation challenge has been extensively studied in recent years. Many works have created new network topologies to handle scale variation implicitly, but are tested with a static inference-time profile. \citet{inception_v1} introduce Inception modules, which contain multiple convolutional layers in parallel, each using different kernel sizes to encode features from various scales. \citet{learning_scale_variant} use an ensemble of single-scale CNNs to create a model that considers both scale-variant and scale-invariant features. Feature Pyramid Network (FPN) and its variants \citep{FPN, AugFPN, MFPN, matrixnets, FP_grids} leverage feature maps from different depths of the convolutional backbone and detect objects of different scales at each level. HRNet \citep{HRNet, HigherHRNet} uses parallel feature streams throughout the convolutional backbone to separate features of different scales. These works all result in a static model at inference-time. Differently, our work looks beyond network topology and considers models that can adapt configurations on the fly to tailor the feature processing for each image.  

Neural architecture search (NAS) algorithms automate network design by training a search algorithm to combine modules from a predefined search space to optimize task performance \citep{NAS_zoph, NAS_baker}. Search algorithms vary from reinforcement learning-based \cite{NAS_zoph, NAS_baker, NASNet, EAS}, to evolutionary algorithms \citep{HNAS, AmoebaNet}, to random search \citep{SMASH} and gradient-based approaches \cite{SNAS}. Despite the innovation in network topology that NAS provides, the resulting networks are ultimately treated as static architectures during inference. 


A limited number of works have explored the concept of making a component of a model dynamic. One approach to handling multi-scale features is to alter the input in an adaptive manner before processing. \citet{dynamic_zoomin} and \citet{when_where_zoom_rl} use a reinforcement learning agent to dynamically select regions of the input to process with high-resolution features, making the detection system more efficient. \citet{learning_to_zoom} introduce a dynamic preprocessing layer to distort inputs based on saliency for a given task to improve spatial sampling. \citet{fine_grained_dynamic_head} frames the feature merging process of an FPN detection head as a routing problem to be learned separately by the model. Deformable convolution \citep{deformable_conv_v1, deformable_conv_v2} is a technique that allows the receptive field for each feature element to be optimized for the detection task. Finally, \citet{switching_CNN} invents a dynamic model which consists of three different backbone networks with different receptive fields and a switch CNN that learns to select the most well-suited backbone for a given image. While our work does not directly result in a dynamic model, we are the first to explore the impressive potential gains from making off-the-shelf models dynamic at inference-time. Before attempting to create and optimize a dynamic model, we feel it is critical to understand the bounds of performance from allowing dynamic behavior. Our comprehensive evaluation process allows us to do this instead of relying on a potentially poorly optimized adaptive module.

\section{Methodology}

Our goal is to show that even vanilla CNN architectures are capable of tremendous recognition performance using an oracle dynamic selection policy. To achieve this, we allow attributes of the network that are commonly static at inference time to be dynamic. The attributes that we study are convolutional kernel stride, kernel dilation, and kernel size. These parameters are of particular interest because they intimately control receptive field growth, spatial fidelity, and in the case of stride and kernel size, computational complexity. These attributes are typically regarded as static properties set by the user during model construction. We implement these dynamic traits such that the pre-trained parameters of the network remain the same, but the way the kernels interface with the input features during convolution changes. By allowing these attributes to be dynamic at inference-time, the network can tailor the feature processing for each input independently, rather than being constrained to the static form used during training.

In this section, we first discuss how kernel stride, dilation, and size affect the feature extraction process, and how they can be made dynamic. We then discuss the motivations for the baseline architectures that we choose to analyze, and how the dynamic layers are placed throughout the models. Next, we explain the comprehensive evaluation process that allows us to analyze the performance bounds of the models assuming different oracle selection policies. Finally, we introduce random option fine-tuning, a method for improving performance volatility.

\subsection{Dynamic attributes}
\textbf{Kernel stride:}
The first attribute that we consider making dynamic is kernel stride. In a 2-dimensional convolutional layer, stride is the parameter that specifies the spatial step size in the convolution operation. Stride is commonly known as the parameter that controls feature down-sampling. In general, for a given layer with an input resolution of $F_H \times F_W$ and stride $S$, the output feature resolution will be $\frac{F_H}{S} \times \frac{F_W}{S}$. Thus, the larger the stride, the more aggressive the down-sampling and the larger the receptive field. For images that contain large scale objects (in pixel-space) or significant secondary contextual features, a larger stride is intuitively more appropriate. Stride is not limited to integer values greater than or equal to one, however. We can also use fractionally-strided convolution to up-sample features. This allows encoding of more detailed spatial information and slows receptive field growth. We implement fractionally-strided convolution using a transposed convolutional layer, where in general the output resolution of a layer with stride $S$ is $S*F_H \times S*F_W$. Note that to implement a fractional-stride of $\frac{1}{2}$ we set $S=2$ for the transposed convolutional layer. See Appendix for more details.

The accepted meta-architecture for CNNs is scale-decreasing, in which feature resolution monotonically decreases throughout the depth of the model. In most modern architectures, setting $S>1$ in a handful of layers is the primary mechanism for down-sampling the features spatially, making them more manageable in memory and expediting receptive field growth. Recently, \citet{spinenet} challenged these heuristics and show that a scale-permuted meta architecture, in which feature resolution can decrease or increase anytime, is superior. Our goal is to study the scale-permuted strategy taken to its extreme by allowing the network to dynamically choose the optimal scale permutation for each input. Most networks exclusively use $S$=1 or $S$=2. We allow strides: \{$\frac{1}{2}$, 1, 2, 3, 4\}. This set of options allows us to test various feature sampling profiles, ranging from up-sampling to extreme down-sampling. Note that we must limit stride options $\frac{1}{2}$ and 1 in certain layers due to memory constraints. For example, in the ResNet-50 model, we do not allow upsampling in the shallow A and B layers.

\textbf{Kernel dilation:}
Dilation is a parameter of a convolutional layer that effectively enlarges the kernel's spatial extent without increasing model parameters. A dilation rate of $D$ inserts $D - 1$ zeros between consecutive filter elements. By expanding the spatial reach of the kernel, we more rapidly grow the receptive field of each spatial feature element without down-sampling the feature resolution. The ability to alter receptive field size has proven useful in multi-scale detection, segmentation, and classification tasks as it has been found that the most appropriate receptive field is directly correlated to the scale of the target object \citep{tridentnet, FPN, DRN, dilation, DetNet, deformable_conv_v1, deformable_conv_v2}. Although many recent works incorporate dilated convolution \citep{tridentnet, dilation, DRN, DetNet}, most current architectures use a default of $D=1$ in all layers. We allow the network to tailor its kernel dilation to each input, rather than attempting to choose an optimal static setting. We allow dilations: \{1, 2, 3, 4, 5\} as we feel this covers a large range of receptive fields from default to well above.

\textbf{Kernel size:}
Kernel size dictates the spatial extent of the filters in a given convolutional layer. The implications of kernel size are similar to dilation in that larger kernel sizes precipitate receptive field growth. Unlike dilation, however, increasing kernel size does not sacrifice feature richness due to sparse spatial sampling. While larger kernel sizes may provide more useful features for certain inputs, they come at the cost of efficiency. Because of factors such as scale and context variation between images, we feel that kernel size is a valid parameter to make dynamic. The optimal choice of kernel size has been experimented with since the conception of CNNs. Most common current meta-architectures \citep{vgg, resnet, resnext, mobilenetv2, densenet} rely on 3$\times$3 layers (\ie $K=3$) to perform pertinent spatial feature extraction.

In our study, we consider making the kernel size of a subset of feature extraction layers (with default size 3$\times$3) dynamic. To make kernel size dynamic without affecting parameter count, we use bilinear spatial interpolation of the kernel tensor during inference. Because the network is trained with static 3$\times$3 layers, we must also correct for the difference in magnitude of the values in the output feature map. To do this, we scale output activations of the dynamic layers by $\alpha = \frac{3^2}{K^2}$. We allow kernel sizes: \{1, 3, 5, 7, 9\}.

\subsection{Dynamic networks}
Most common current network designs consist of four main stages where feature depth and resolution is kept consistent. In the first module of each stage, there typically exists a transition layer that increases the channel dimension and reduces feature resolution. For example, ResNets are comprised of a stem followed by four stages of Bottleneck modules. Feature depth and resolution is reduced in the first Bottleneck of each stage. To study the effect of dynamic layers across model depth without an evaluation that is too expensive, we choose four locations distributed throughout the model to replace with dynamic variants. In ResNet, we allow the 3$\times$3 layer in the first Bottleneck of each stage to be dynamic. We refer to these four locations as A, B, C, and D, respectively. Note that we refer to a model with all four dynamic layers present as the ABCD variant.

To fairly evaluate the potential of dynamic CNNs we also consider a variety of additional architectures. We examine MobileNetv2 \citep{mobilenetv2} to test dynamic depth-wise separable convolution, DenseNet-121 \citep{densenet} to test the effect of dynamic convolution with feature concatenation, and ResNeXt-50 \citep{resnext} to test a dynamic version of the split-transform-aggregate paradigm. The placement of dynamic layers in these networks is similar in reason to the dynamic ResNet variant outlined above: place dynamic layers where the default down-sampling occurs. For more information on dynamic model layout, see the Appendix. For all of the architectures considered, we download pre-trained parameters from the PyTorch Model Zoo \citep{pytorch}. 


\subsection{Comprehensive evaluation}
Recall that the purpose of this work is to study the true potential of ITD models. To do so, we consider the performance of an ``oracle" attribute selector, which is presumed to select the most optimal parameter configurations (\ie permutations) for every test input. To realize this oracle selector, we use a comprehensive evaluation procedure in which we run each sample through all possible model configurations and track best-case top-1 accuracy. The best-case performance across all option permutations for each sample tells us how an ITD model would perform with an optimal selector. For each sample $(x_i, y_i) \in \mathcal{D}_{val}$, where $x_i$ is the image, $y_i$ is the label, and $\mathcal{D}_{val}$ is the ImageNet validation set, we collect a sequence of output vectors $O_i = \left \{ \mathrm{softmax}(f(x_i, v_m, \theta)) \right \}_{m=1}^{M}$ and class predictions $P_i = \left \{ \mathrm{arg}\max_y{\left (O_i^m\left [ y \right ]  \right )} \right \}_{m=1}^{M}$ by forwarding $x_i$ through model $f$ parameterized by an attribute configuration $v_m$ (\ie a stride, dilation, or size permutation) and pre-trained parameters $\theta$, for all $M$ possible permutations. We want to sort each of the $M$ output vectors by their quality or ``correctness", which we quantify as the confidence in the true label class, $t_i^m = O_i^m\left [ y = y_i \right ]$. However, the best-case prediction vector is not necessarily the one with the highest confidence in the true class. To account for this, we add 1 to $t_i^m$ if the $m^{\mathrm{th}}$ output correctly predicts the target class. Thus, the quality of the $m^{\mathrm{th}}$ configuration is:
\vspace{-2mm}
\begin{equation} \label{eq:quality}
q_i^m = \begin{cases}
 & t_i^m + 1 \text{\quad if } \mathrm{arg}\max{\left (O_i^m  \right )}=y_i \\ 
 & t_i^m \text{\quad\quad\ \  else } 
\end{cases}
\end{equation}
and we combine these into $Q_i = \left \{ q_i^m \right \}_{m=1}^{M}$. In this work, we refer to the permutation that yields the highest quality score for a given sample is the model's ``preferred" configuration for that sample. Using these quality scores, we compute the best, median, and worst-case accuracy using the following prediction rules for each sample:
$p_{\mathrm{best}} = P_i^{m=\mathrm{arg}\max{\left ( Q_i \right )}}$,
$p_{\mathrm{median}} = P_i^{m=\mathrm{arg}\mathrm{\ median}{\left ( Q_i \right )}}$,
$p_{\mathrm{worst}} = P_i^{m=\mathrm{arg}\min{\left ( Q_i \right )}}$.

\subsection{Random option fine-tuning}
While our hypothesis is that off-the-shelf pre-trained models are capable of very high accuracy assuming a best-case selection policy, we also acknowledge that a sub-optimal selection policy could yield poor performance. We attempt to improve this lower performance bound using a lightweight fine-tuning stage. We fine-tune the pre-trained model for 15 epochs on the ImageNet training set where for each batch, we randomly select an option permutation and configure the dynamic layers accordingly before the forward pass. By performing this Random Option Fine-tuning (ROF) stage, the model can learn to become more robust to feature scale variation that it did not encounter during standard training. While this may not significantly improve the best-case performance, it reduces performance volatility, which we define as the difference between best-case and worst-case accuracy. Reducing volatility will be important for actually implementing an adaptive dynamic model, where the selection policy will likely be sub-optimal.

\section{Experiments}

We first discuss the capabilities of dynamic variants of common off-the-shelf models in section \ref{sec:performance_bounds}. In section \ref{sec:distribution_of_preds} we investigate the distribution of task predictions across permutations. Section \ref{sec:layer_choice} covers the effect of dynamic layer choice, and section \ref{sec:combining_attributes} explores the effect of combining dynamic attributes in the same layers. Section \ref{sec:examining_model_preferences} contains a study on model preferences, and how characteristics of the input images affect the best attribute options. Finally, in section \ref{sec:efficiency}, we explore the potential efficiency gains that ITD models can provide.

\begin{table}[t]
\centering
\renewcommand\arraystretch{1.2}
\resizebox{\linewidth}{!}{
\begin{tabular}{l|c|ccc|ccc|ccc}
\toprule
\multicolumn{1}{c|}{Model} & Static & \multicolumn{3}{c|}{Dyn. Stride} & \multicolumn{3}{c|}{Dyn. Dilation} & \multicolumn{3}{c}{Dyn. Size} \\ \hline
                           &        & W        & M         & B         & W          & M         & B         & W      & M         & B        \\ \cline{3-11} 
ResNet-50                  & 76.5   & 0.1      & 11.2      & 93.7      & 21.6       & 58.0      & 88.9      & 0      & 32.6      & 90.7     \\
ResNeXt-50                 & 77.0   & 0        & 1.8       & 94.0      & 0.2        & 4.8       & 90.8      & 0      & 6.2       & 92.4     \\
MobileNetV2                & 71.9   & 0        & 0.4       & 92.1      & 0.0        & 0.5       & 83.7      & 0      & 0.1       & 82.3     \\
DenseNet-121               & 74.5   & 0        & 14.1      & 93.2      & 18.9       & 55.8      & 87.5      & 0      & 27.7      & 88.5   \\
\bottomrule
\end{tabular}
}
\vspace{-2mm}
\caption{\textbf{Performance bounds for common architectures.} A comparison of the worst (W), median (M) and best-case (B) accuracy (\%) for common dynamic models (ABCD).}
\label{tab:all_achitectures}
\end{table}
\begin{table}[t]
\centering
\renewcommand\arraystretch{1.2}
\resizebox{\linewidth}{!}{
\begin{tabular}{c|ccc|ccc|ccc}
\toprule
Model                   & \multicolumn{3}{c|}{Dyn. Stride}             & \multicolumn{3}{c|}{Dyn. Dilation}            & \multicolumn{3}{c}{Dyn. Size}                 \\ \hline
                        & W            & M             & B             & W             & M             & B             & W             & M             & B             \\ \cline{2-10} 
ResNet-50 (Pre-trained) & 0.1          & 11.2          & \textbf{93.7} & 21.6          & 58.0          & \textbf{88.9} & 0             & 32.6          & \textbf{90.7} \\
ResNet-50 (After ROF)   & \textbf{1.5} & \textbf{57.6} & 92.2          & \textbf{64.6} & \textbf{76.0} & 84.6          & \textbf{49.6} & \textbf{75.3} & 88.1    \\
\bottomrule
\end{tabular}
}
\vspace{-2mm}
\caption{\textbf{Effect of ROF.} A comparison of the worst (W), median (M), and best-case (B) accuracy of the dynamic ResNet-50 model (ABCD) before and after an ROF stage.}
\vspace{-4mm}
\label{tab:after_rf}
\end{table}

\subsection{Performance bounds for common models}
\label{sec:performance_bounds}
The first experiment that we conduct is an analysis of the worst, median, and best-case top-1 accuracy on ImageNet for four popular architectures, each with four dynamic layers (\ie ABCD models). From Table \ref{tab:all_achitectures}, it is clear that ITD convolution is a very powerful technique. On average across the four architectures, the best-case accuracy improves by 18.3\%, 12.8\%, and 13.5\% with dynamic stride, dilation, and size, respectively. Also, note that the stride-dynamic ResNet-50 is capable of 93.7\% accuracy with a best-case selection policy. This level of performance surpasses modern state-of-the-art NAS models that have over 20x more parameters \cite{efficientnet, NFNet}! 

One notable challenge with the dynamic approach is the large volatility between worst- and best-case performance. A simple way to combat this is to perform an inexpensive ROF step to the pre-trained model. In Table \ref{tab:after_rf}, we show that ROF significantly improves worst-case and median-case accuracy regardless of dynamic attribute. Across the three dynamic attributes, worst-case accuracy improves by an average of 31.3\% and median accuracy improves by 35.7\%. This trend shows the ability of ROF to make models robust to feature scale variation. The slight reduction in best-case accuracy after ROF can be attributed to a narrowing of the prediction distribution (see Section \ref{sec:distribution_of_preds}).

\subsection{Distribution of predictions} 
\label{sec:distribution_of_preds}
\begin{figure}[t]
    \centering
    \includegraphics[width=\columnwidth]{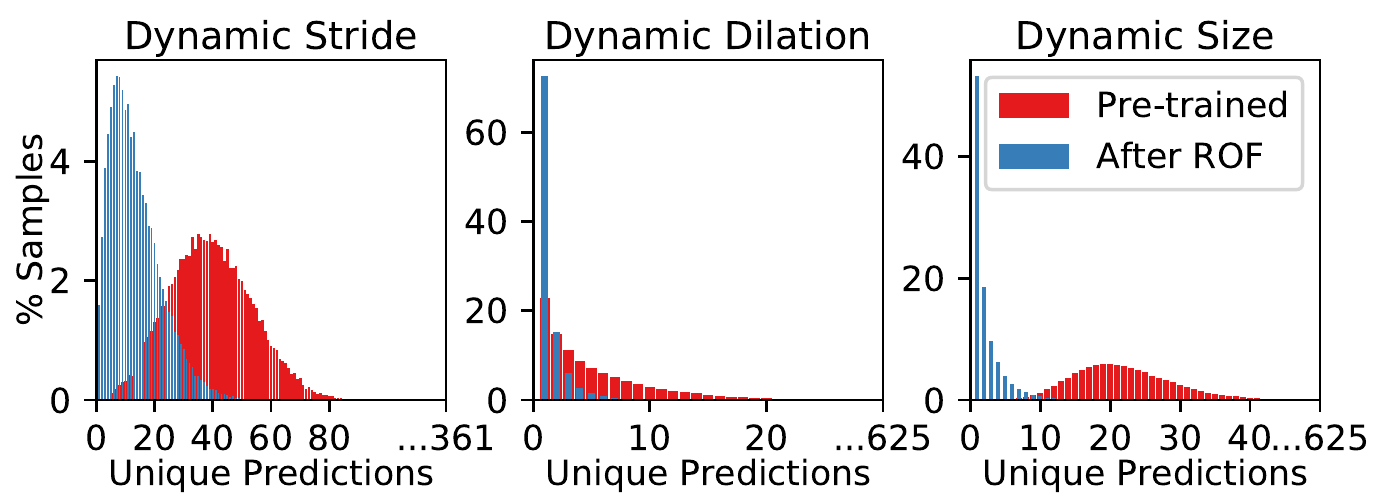}
    \vspace{-6mm}
    \caption{\textbf{Histograms of unique predictions.} For each dynamic component, we visualize the number of unique predictions made by all option permutations for each sample by the ResNet-50 (ABCD) model.}
    \label{fig:unique_hist}
    \vspace{-4mm}
\end{figure}

While the raw upper bounds of performance are impressive, it is useful to know how the task predictions resulting from differently configured models are distributed. One way to visualize this is in Figure \ref{fig:unique_hist}, where we plot histograms showing the number of unique predictions made across all dynamic option permutations for each attribute. Notice that despite the fact that there are 361, 625, and 625 permutations for a ResNet-50 (ABCD) with dynamic stride, dilation, and size, respectively, a significant proportion of them result in similar predictions on most samples. After we perform an ROF stage, we see that the distribution of unique predictions becomes even more narrow, resembling the shape of an exponential distribution for models with dynamic dilation and size. In fact, after ROF on the dynamic dilation model, 73\% of samples are predicted as the same class by every configuration. The difference in unique prediction distributions before and after ROF helps to explain the performance bound shifts from Section \ref{sec:performance_bounds}. After ROF, the dynamic models are more robust to feature scale variation, meaning that models configured differently tend to make more similar predictions. This reduces performance volatility, but since we are casting a smaller net of unique predictions, the upper bound performance drops slightly.

\begin{figure}[t]
    \centering
    \includegraphics[width=0.9\columnwidth]{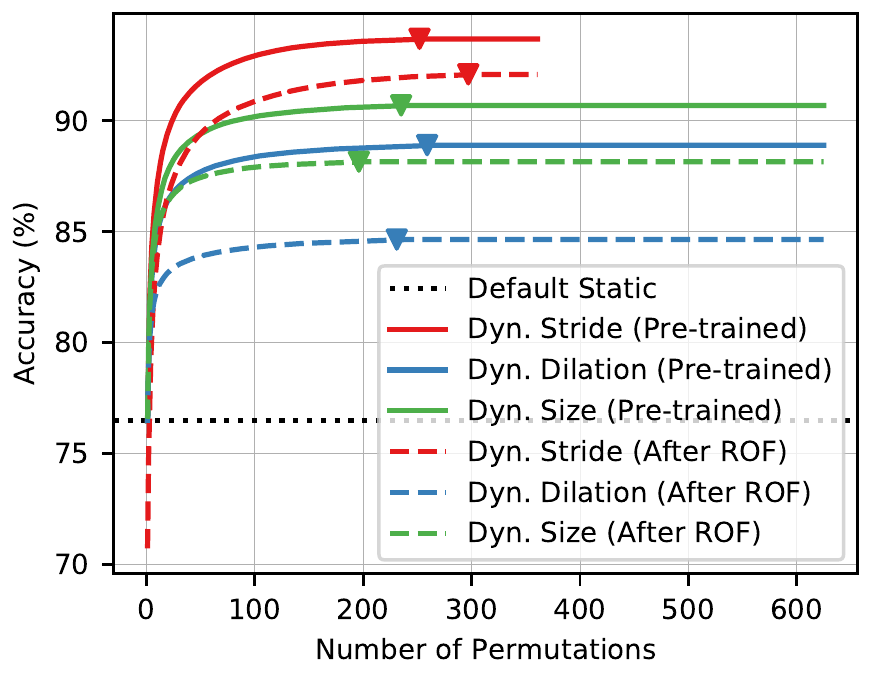}
    \vspace{-3mm}
    \caption{\textbf{Number of option permutations vs. accuracy.} Using a greedy aggregation algorithm, we compute the best accuracy possible for a given number of option permutations for ResNet-50 (ABCD). We visualize the relationship between \# permutations and best-case accuracy. Triangle markers indicate the minimum permutations required to achieve the maximum best-case accuracy.}
    \label{fig:numperms_vs_bestacc}
    \vspace{-4mm}
\end{figure}

\begin{figure*}[t]
    \centering
    \includegraphics[width=1.9\columnwidth]{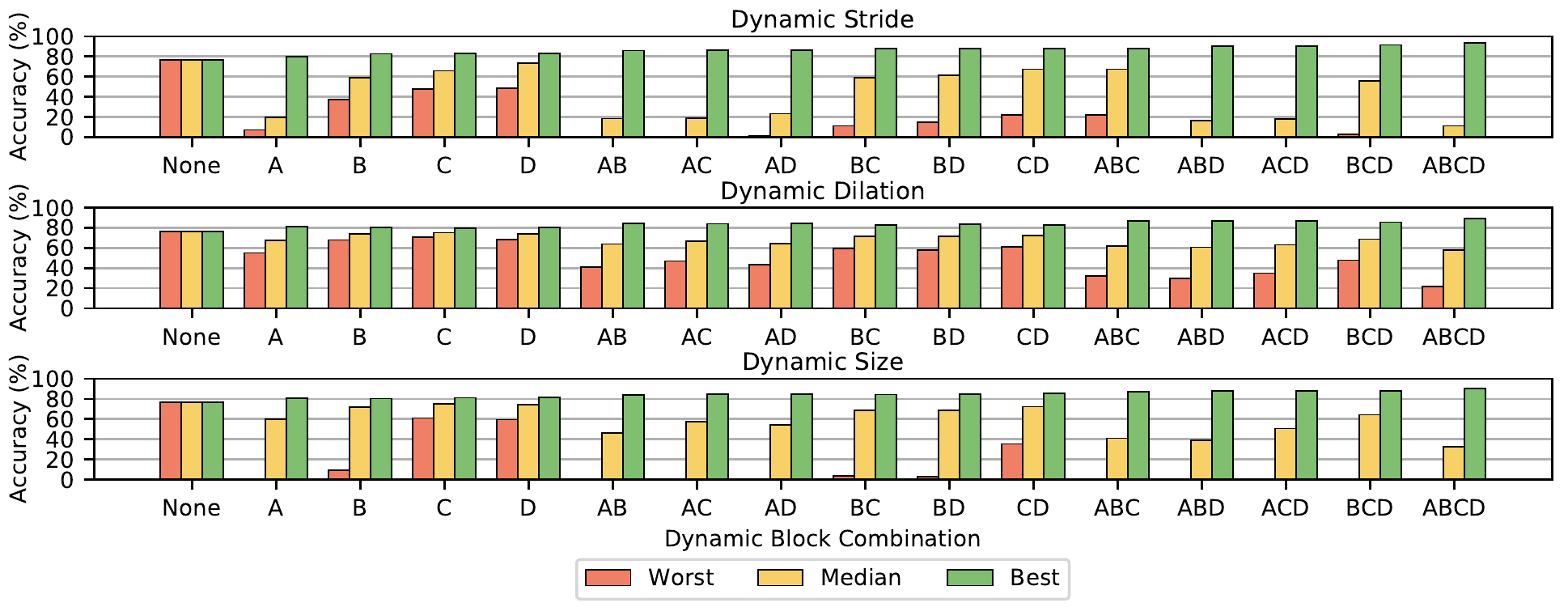}
    \vspace{-3mm}
    \caption{\textbf{Effect of dynamic layer location.} For each dynamic attribute we compare the performance bounds for every combination of dynamic layer locations in a ResNet-50 pre-trained model.}
    \label{fig:layerwise_acc}
    \vspace{-4mm}
\end{figure*}

Now that we are aware of the significant duplication of predictions from different configurations, a natural question is: do we need \textit{all} possible permutations tested in the comprehensive evaluation process to achieve substantial performance improvements? Also, what is the trade-off between the number of allowed permutations and the best-case accuracy? To answer these questions, we combine our comprehensive evaluation procedure with a greedy accumulation algorithm to iteratively compute the best-case accuracy for subsets of one to all permutations. The resulting trade-off curves are shown in Figure \ref{fig:numperms_vs_bestacc}. The near vertical early growth in all the curves indicates that we really only need a handful of option permutations to achieve performance far superior to the baseline static architecture. For example, the pre-trained dynamic stride model requires just six permutations to reach 85\% accuracy. This result is particularly useful for future works that attempt to leverage the benefits of ITD models, as it shows that we can significantly trim the number of configurations to consider without sacrificing the performance ceiling.

\subsection{Effect of dynamic layer choice}
\label{sec:layer_choice}
In this experiment we seek to understand the most influential dynamic layer locations in a ResNet-50 pre-trained model. In Figure \ref{fig:layerwise_acc} we see that the more dynamic layers we include, the higher the best-case performance. This improved upper bound also comes with higher volatility, however. Although this trend can be partially explained by the higher volume of predictions used in the comprehensive evaluation, it demonstrates the advantage of diversifying dynamic layer placement. Another pattern to note is that in general, the deeper the dynamic layers are placed, the less volatile the performance is. We argue that this is due to the fact that CNNs are a sequential composition of functions, so choosing an inadequate configuration for a shallow layer has a more profound impact on the total feature extraction than a deep layer. Another notable finding is that even with a single dynamic layer and attribute, a pre-trained ResNet-50 model is capable of over 80\% accuracy on ImageNet. Finally, even though we limit our study to four dynamic layers, we can safely assume that adding more would improve the best-case accuracy further. However, when it comes to to implementing a true adaptive dynamic model, having fewer total option permutations may be advantageous.

\begin{figure*}[t]
    \centering
    \includegraphics[width=1.9\columnwidth]{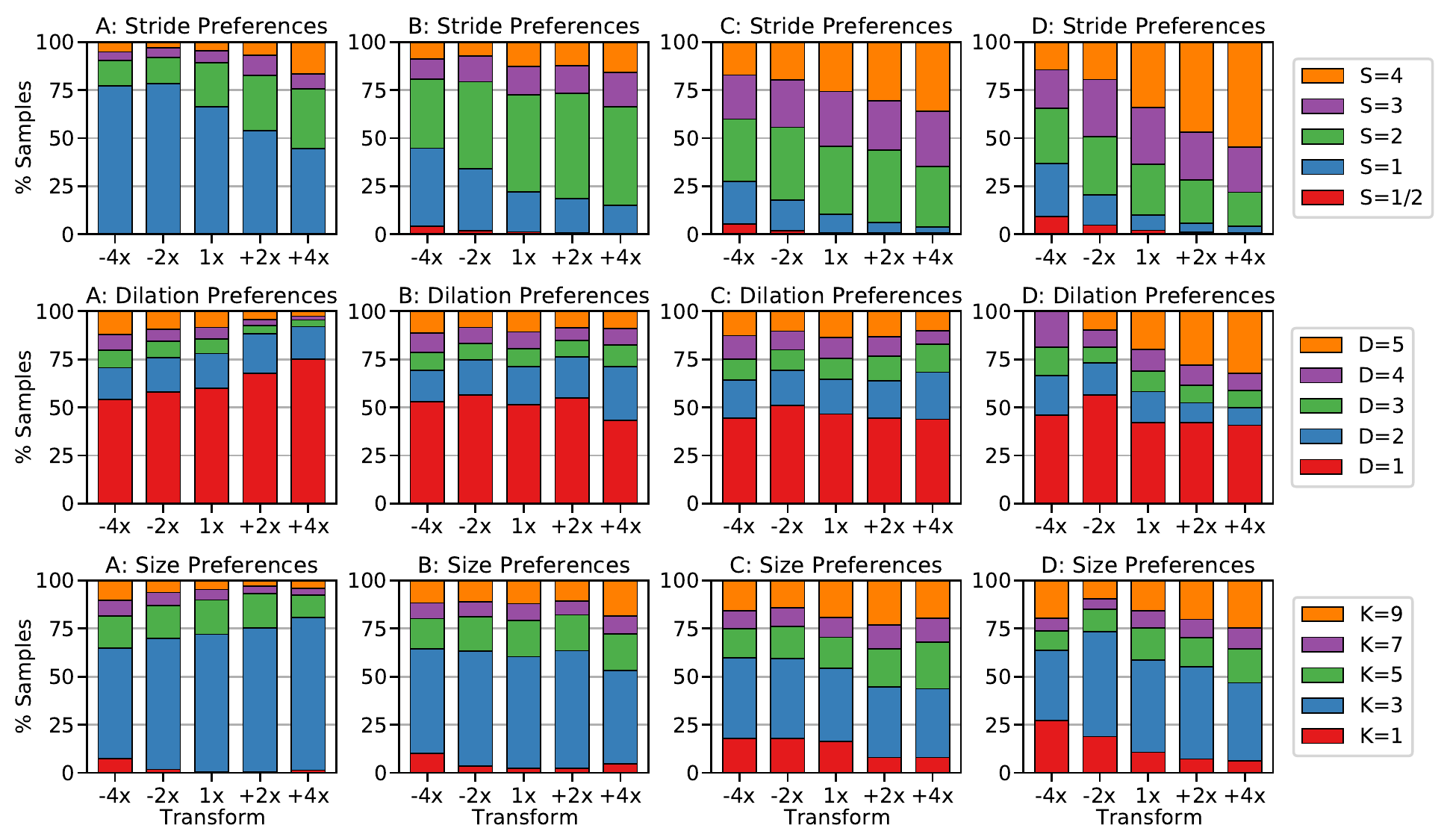}
    \vspace{-3mm}
    \caption{\textbf{The effect of feature scale on model preference.} The distribution of model preference is compared for each dynamic layer when a series of scale transforms is applied to each input image. Note that the transform +2x refers to resizing the height and width of each input to 2x its original size of 224$\times$224 (\ie 448$\times$448). The three rows correspond to ResNet-50 stride-, dilation-, and size-dynamic models, respectively.}
    \label{fig:explain_scale}
    \vspace{-4mm}
\end{figure*}

\subsection{Combining dynamic attributes}
\label{sec:combining_attributes}
\begin{table}[t]
\centering
\renewcommand\arraystretch{1.2}
\resizebox{0.9\linewidth}{!}{
\begin{tabular}{ccc|cc|ccc}
\toprule
\multicolumn{3}{c|}{Dyn. Attribute} & \multicolumn{2}{c|}{\# Permutations} & \multicolumn{3}{c}{Top-1 Acc. (\%)}            \\ \hline
Stride     & Dilation     & Size    & Per Attr.           & Total          & W             & M             & B             \\ \hline
\ding{51}          &              &         & 361                 & 361            & 0.1           & 11.2          & 93.7          \\
           & \ding{51}            &         & 625                 & 625            & \textbf{21.6} & \textbf{58.0} & 88.9          \\
           &              & \ding{51}       & 625                 & 625            & 0             & 32.6          & 90.7          \\
\ding{51}          & \ding{51}            &         & 25                  & 625            & 0.4           & 47.9          & 95.0          \\
\ding{51}          &              & \ding{51}       & 25                  & 625            & 0             & 35.3          & \textbf{95.8} \\
           & \ding{51}            & \ding{51}       & 25                  & 625            & 0.2           & 43.0          & 92.7          \\
\ding{51}          & \ding{51}            & \ding{51}       & 8                   & 512            & 0.1           & 32.9          & 94.7 \\
\bottomrule
\end{tabular}
}
\vspace{-2mm}
\caption{\textbf{Comparison of dynamic attributes.} A comparison of the worst (W), median (M), and best-case (B) accuracy of a dynamic ResNet-50 model (ABCD) with different combinations of dynamic attributes.}
\vspace{-2mm}
\label{tab:attribute_compare}
\end{table}
In the previous experiments, we compare the effectiveness of dynamic kernel stride, dilation, and size independently. Until this point, we have ignored the possibility of combining different dynamic attributes in the same layers. A simple way of analyzing the bounds of multi-attribute dynamic models would be to perform a comprehensive evaluation with the cartesian product of option permutations from each attribute. However, this approach is not only incredibly expensive but it would lead to unfair comparisons against single-dynamic-attribute models. Both drawbacks are due to the polynomial increase in possible permutations. For example, an ABCD model with all three dynamic attributes would have \num{1.4e8} permutations. With this many permutations, and corresponding predictions, we are bound to achieve a better upper bound from prediction entropy alone. To make a fair comparison, we choose 625 (the default permutation count for dilation and size attributes) to be the maximum total permutations allowed. In other words, we allow $R$ permutations per attribute following: 
\begin{equation} \label{eq:R_choice}
\max{(R)}\ \  \mathrm{s.t.}\ \  R^{\mathrm{\#perms}} < 625.
\end{equation}
We sample the $R$ most important permutations per attribute using the greedy aggregation algorithm from section \ref{sec:distribution_of_preds}. 

The results of this experiment are in Table \ref{tab:attribute_compare}. Notice that in all cases, models with combined dynamic attributes yield higher best-case accuracy than any of the included single attribute models. This finding indicates that the benefits of different dynamic attributes are not mutually exclusive, even though they are all essential parameters to control receptive field. We find that given an allowance of 625 permutations, the combination of stride and kernel size yields the largest best-case accuracy, 95.8\%. We also note that the worst-case accuracy reduces to close to zero when we allow combined dynamic attributes, but this can be improved with an ROF stage as shown in section \ref{sec:performance_bounds}.

\subsection{Examining model preferences}
\label{sec:examining_model_preferences}
A dynamic model has an optimal configuration for every input. In this section, we investigate the distribution of best-case option selections (\ie preferences), as well as attempt to correlate model preferences with characteristics of input images. The first experiment that we conduct relates to feature scale. A common belief is that the larger the input features, the larger the preferred receptive field will be. To test this, we resize the standard ImageNet validation images to different resolutions ranging from 4x down-sample to 4x up-sample and observe the distributions of option preferences in each layer of a ResNet-50 (ABCD) pre-trained model. Note that we first resize the raw images to 256$\times$256, then center-crop to 224$\times$224 before resizing again for the experiment, so each image version has identical contextual features. Figure \ref{fig:explain_scale} contains the full results. 

Notice all four layers of the stride-dynamic model (top row) show the expected trend. As image scale increases, the preference of the largest stride ($S$=4) increases while the preference for $S$=1 decreases. In the dilation-dynamic model, we observe a different behavior. As input scale increases, the shallow A layer prefers the smaller dilation options. Note that this is the inverse of what we expect. In the middle B and C layers, there does not exist an obvious relationship between feature scale and option preference. In the D layer, however, we see more of the expected relationship as evidenced by the increased use of $D$=5 as input scale increases. The size-dynamic model shows very similar preferences to the dilation-dynamic model: inverse of expected relationship in the shallow A layer, and expected relationship in the deep D layer.

Another characteristic of images that has been correlated to architecture design is global context in images. The intuition is that the more contextual information that surrounds the object of interest, the larger the preferred receptive field will be. This intuition is based on the fact that like humans, classifiers use features surrounding the target object as context clues when determining what the object is. The more contextual information we have available, the more reasoned our prediction will be. To test this we use a similar experiment to the one above, but instead of studying how preference distributions change across image scales we look at how preference distributions change across different levels of context. We change context by resizing the standard validation images to a fixed 256$\times$256, then cropping different areas around the image center ranging from 50$\times$50 to 250$\times$250 pixels. The results of this test are shown in Figure \ref{fig:explain_context} in the Appendix. In general, the relationships between global context and option preference are nearly identical across attributes and layers to the relationships regarding image scale. These results confirm that feature scale and context are indeed linked to optimal model configuration.

Our main takeaway from these interpretability studies is that although the relationship between feature scale, global context, and optimal receptive field has been studied and leveraged by many works \citep{inception_v1, tridentnet, DRN, HRNet, SSD, DSSD}, it is not as straightforward as it is made out to be. Using scalar receptive field as a metric for determining what the optimal model configuration should be is not adequate. First, scalar receptive field is a misleading metric to begin with. Models with very different configurations across their depths can share a similar receptive field in the final feature map. We propose that receptive field \textit{growth} is a more explanatory metric. Receptive field growth elucidates the receptive field acceleration across the depth of the model, which, as we have seen, is critical for optimizing the dilation- and size-dynamic models. Also, the attribute used for altering the receptive field affects the preferences of the model. This is clearly apparent in Figure \ref{fig:explain_scale}, which shows that models with dynamic dilation and size have an inverse scale-to-preferred-option relationship in the shallow layers, while the model with dynamic stride has more intuitive preferences across all layers.

\subsection{Efficiency benefits}
\label{sec:efficiency}
\begin{figure}[t]
    \centering
    \includegraphics[width=0.85\columnwidth]{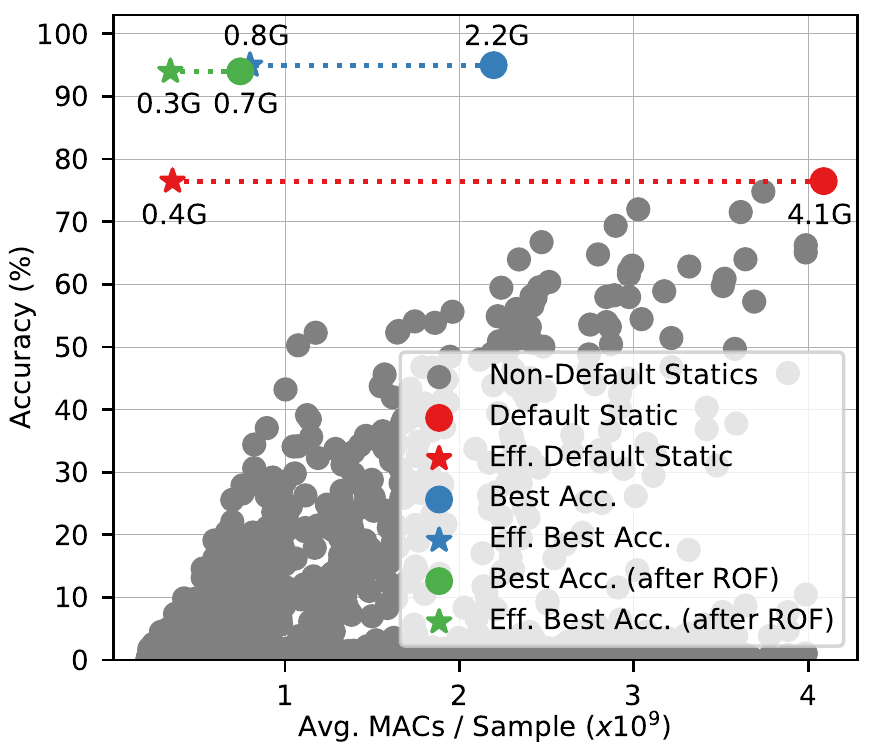}
    \vspace{-2mm}
    \caption{\textbf{Accuracy vs. efficiency trade-off.} A visualization of the accuracy vs. efficiency trade-off of static models, best-case dynamic models, and their efficient variants. Points are labeled with the average MACs / sample. A pre-trained ResNet-50 (ABCD) dynamic model is considered.}
    \label{fig:efficiency}
    \vspace{-2mm}
\end{figure}

By now we have established the immense potential of dynamic models for improving accuracy, but dynamic models can be made incredibly efficient as well. The reason for this efficiency benefit is twofold: (1) reducing the spatial extent of a feature map by increasing stride means the downstream layers will necessarily incur less MACs per sample, and (2) interpolating the default 3$\times$3 kernel to 1$\times$1 also reduces MACs in that layer. Thus, over a test set, the average efficiency per sample can be reduced. To explore the efficiency potential, we consider a ResNet-50 (ABCD) model and allow strides: \{1, 2, 3, 4\} and sizes: \{1, 3\}. To establish an upper bound of efficiency for a given accuracy, we use an exhaustive evaluation that finds the most efficient option permutation that yields the same prediction as the model that we are emulating for every sample.

Figure \ref{fig:efficiency} shows the accuracy vs. efficiency trade-off of static models, best-case (accuracy) dynamic configurations, and their efficient variants. The gray markers show the performance of the static variants of every possible option permutation. The red circle marker shows the performance of the default static ResNet-50 configuration. We find that we can configure a dynamic variant (red star marker) to achieve the same 76.5\% accuracy while reducing computational cost by over 10x! We can also improve the efficiency of the pre-trained best-case accuracy configuration by 2.75x. Finally, we note that the efficient variant of the best-case configuration after ROF achieves 94.0\% accuracy at an average of 0.3 GMACs per sample, making this dynamic ResNet-50 as efficient as the lightweight MobileNetV2 \citep{mobilenetv2}.

Most existing works regarding network efficiency attempt to reduce parameter size in memory. This is often done by weight pruning, quantization, and clever encoding schemes like Huffman coding \citep{deep_compression, compression_survey}. ITD models offer a different approach to gaining efficiency that in no way affects the parameters themselves. This difference may imply that the benefits from dynamic models and pruning/quantization are independent, meaning their combination could yield a compounded improvement. This investigation is left for future work. 

\section{Conclusion}

In this paper, we have exposed the exciting potential of ITD models. When configured properly, even off-the-shelf models have the potential to greatly exceed the accuracy and efficiency levels of their static variants. Techniques such as ROF can be used to improve the performance bounds by reducing performance volatility and requiring fewer options to choose from. We hope that this inspires future work that leverages this untapped potential of dynamic convolution. We believe that allowing networks to tailor their configuration to each input is an essential step in advancing state-of-the-art performance across the vision domain.

{\small
\bibliographystyle{plainnat}
\bibliography{egbib}
}

\clearpage
\onecolumn
\section*{Appendix}

\subsection*{Implementation details}
\paragraph{Fractionally-strided convolution:}
An important detail when implementing a fractionally-strided convolution with a transposed convolution in PyTorch is to flip the kernel spatially about the center, and swap the input and output channel dimensions of the kernel tensor before performing the convolution. These modifications are necessary because a transposed convolution is essentially implemented as a standard convolution with a flipped kernel striding over the output feature rather than the input.

\paragraph{Random option fine-tuning:}
In this work, we fine-tune for 15 epochs. For the dynamic ResNet-50 model, we use SGD and start with a learning rate of 0.001 and decay it to 0.0001 after 10 epochs. We also set momentum to 0.9 and weight decay to 0.0001.

\paragraph{Dynamic layer locations:}
For the ResNet-50 model, we make the 3$\times$3 convolutional layer in the first Bottleneck module in stages \textit{conv2}, \textit{conv3}, \textit{conv4}, and \textit{conv5} dynamic. ResNeXt-50 is very similar, but the layers that we modify are group convolutions. In MobileNetV2, we allow the 3$\times$3 depth-wise convolutions in all four Bottlenecks that have a default stride of 2 to be dynamic. DenseNet-121 is slightly different because it does not use strided convolution to down-sample feature resolution. Rather, it uses three strided pooling layers (\ie \textit{Transition} layers). Thus, we implement the dynamic stride by changing the stride of the average pool layers in the three \textit{Transition} layers as well as the stride in the max pool layer in the stem. We implement dynamic dilation and kernel size in the first \textit{Dense Layer} in each of the four \textit{Dense Blocks}.

\subsection*{Additional figures}
\begin{figure*}[ht]
    \centering
    \includegraphics[width=0.9\columnwidth]{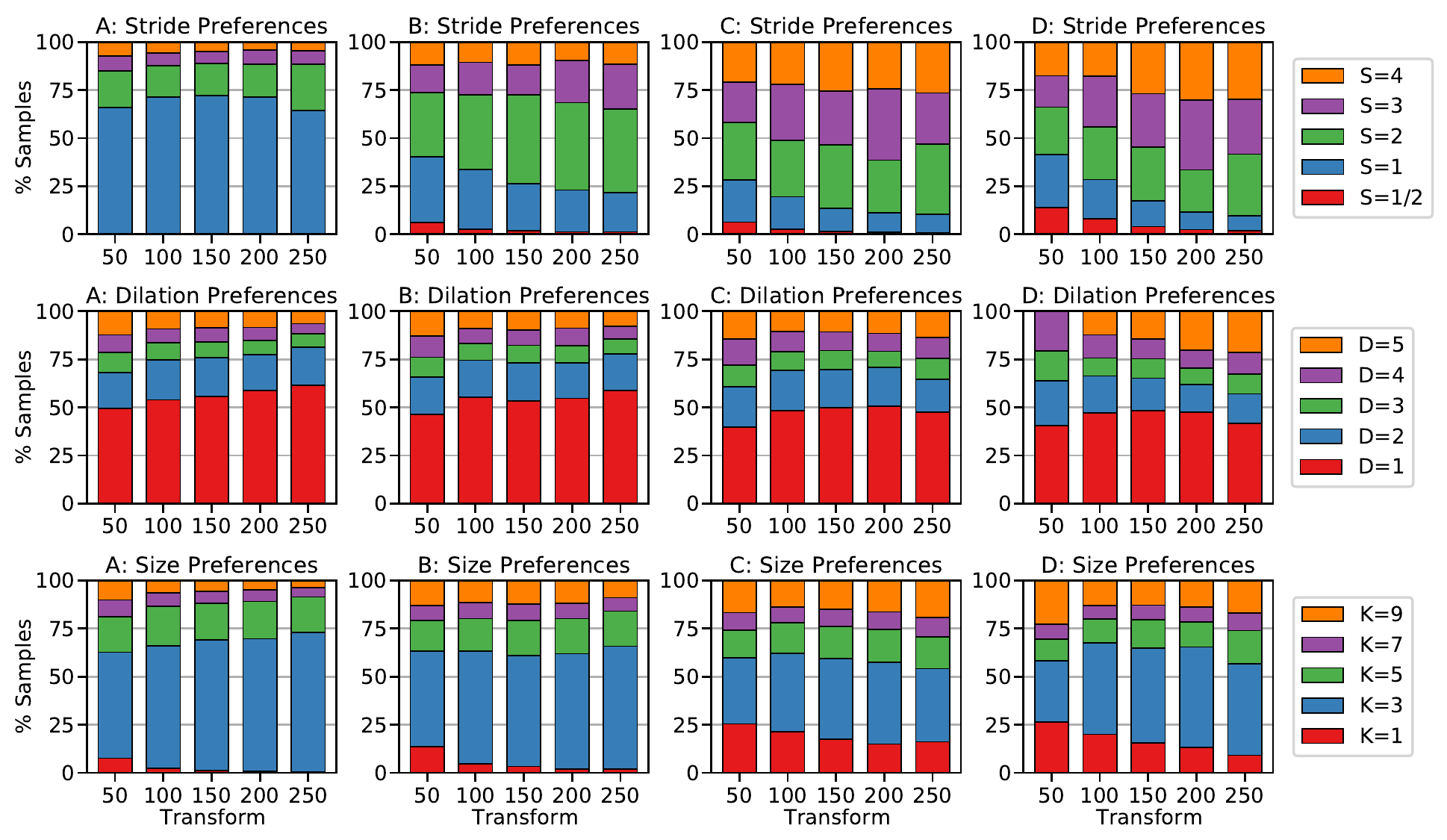}
    \caption{\textbf{The effect of global context on model preference.} The distribution of model preference is compared for each dynamic layer over a series of crop transforms to each input image. Note that the transform 200 refers to center-cropping a 200$\times$200 region of the resized input. The three rows correspond to ResNet-50 stride-, dilation-, and size-dynamic models, respectively.}
    \label{fig:explain_context}
    \vspace{-10pt}
\end{figure*}

\begin{figure*}[ht]
    \centering
    \includegraphics[width=1\columnwidth]{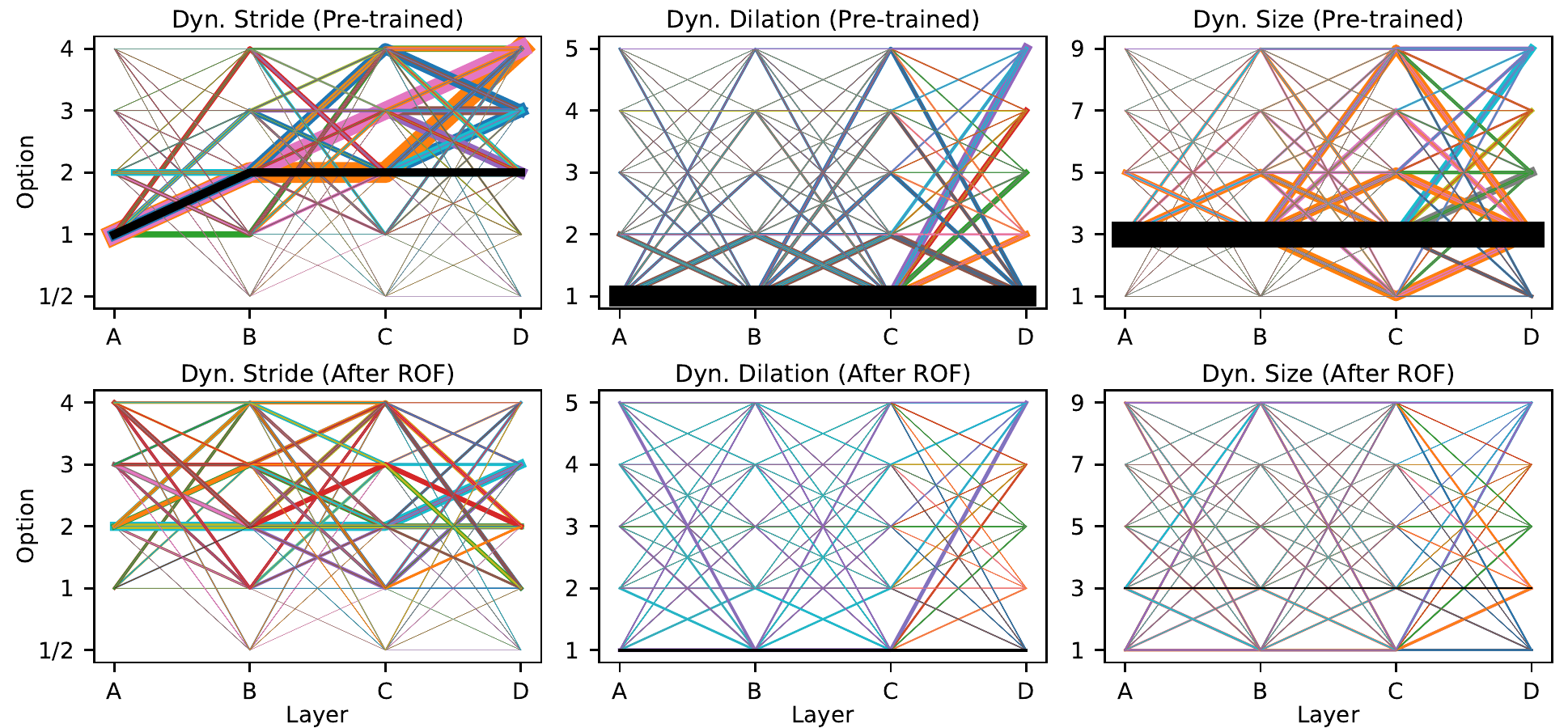}
    \caption{\textbf{Preference paths of ResNet-50 (ABCD).} Similar to Figure \ref{fig:best_paths}, this plot is an intuitive visualization of the option permutation preferences of a ResNet-50 (ABCD) model. Each different colored line stretches from layer A to D and represents a different possible option sequence over the depth of the model. Line thickness encodes the proportion of the validation samples that most prefer that permutation of options. The lines corresponding to the default permutations are colored black.}
    \label{fig:all_best_paths}
    \vspace{-10pt}
\end{figure*}

\end{document}